\documentclass[conference]{IEEEtran}
\IEEEoverridecommandlockouts
\usepackage{cite}
\usepackage{amsmath,amssymb,amsfonts}
\usepackage{algorithmic}
\usepackage{graphicx}
\usepackage{textcomp}
\usepackage{xcolor}
\def\BibTeX{{\rm B\kern-.05em{\sc i\kern-.025em b}\kern-.08em
    T\kern-.1667em\lower.7ex\hbox{E}\kern-.125emX}}
\begin{document}

\title{KVL-BERT: Knowledge Enhanced Visual-and-Linguistic BERT for Visual Commonsense Reasoning}

\author{\IEEEauthorblockN{1\textsuperscript{st} Dandan Song}
\IEEEauthorblockA{\textit{Beijing Institute of Technology} \\
Beijing 100081, China \\
sdd@bit.edu.cn}
\and
\IEEEauthorblockN{2\textsuperscript{nd} Siyi Ma}
\IEEEauthorblockA{\textit{Beijing Institute of Technology} \\
Beijing 100081, China \\
masiyi@bit.edu.cn}
\and
\IEEEauthorblockN{3\textsuperscript{rd} Zhanchen Sun}
\IEEEauthorblockA{\textit{Beijing Institute of Technology} \\
Beijing 100081, China \\
sunzhanchen@bit.edu.cn}
\and
\IEEEauthorblockN{4\textsuperscript{th} Sicheng Yang}
\IEEEauthorblockA{\textit{Beijing Institute of Technology} \\
Beijing 100081, China \\
yangsicheng@bit.edu.cn}
\and
\IEEEauthorblockN{5\textsuperscript{th} Lejian Liao}
\IEEEauthorblockA{\textit{Beijing Institute of Technology} \\
Beijing 100081, China \\
liaolj@bit.edu.cn}
}

\maketitle

\begin{abstract}
Reasoning is a critical ability towards complete visual understanding. To develop machine with cognition-level visual understanding and reasoning abilities, the visual commonsense reasoning (VCR) task has been introduced. In VCR, given a challenging question about an image, a machine must answer correctly and then provide a rationale justifying its answer. The methods adopting the powerful BERT model as the backbone for learning joint representation of image content and natural language have shown promising improvements on VCR. However, none of the existing methods have utilized commonsense knowledge in visual commonsense reasoning, which we believe will be greatly helpful in this task. With the support of commonsense knowledge, complex questions even if the required information is not depicted in the image can be answered with cognitive reasoning. Therefore, we incorporate commonsense knowledge into the cross-modal BERT, and propose a novel Knowledge Enhanced Visual-and-Linguistic BERT (KVL-BERT for short) model. Besides taking visual and linguistic contents as input, external commonsense knowledge extracted from ConceptNet is integrated into the multi-layer Transformer. In order to reserve the structural information and semantic representation of the original sentence, we propose using relative position embedding and mask-self-attention to weaken the effect between the injected commonsense knowledge and other unrelated components in the input sequence. Compared to other task-specific models and general task-agnostic pre-training models, our KVL-BERT outperforms them by a large margin.
\end{abstract}


\begin{figure*}[t]
  \centering
  \includegraphics[width=0.85\linewidth]{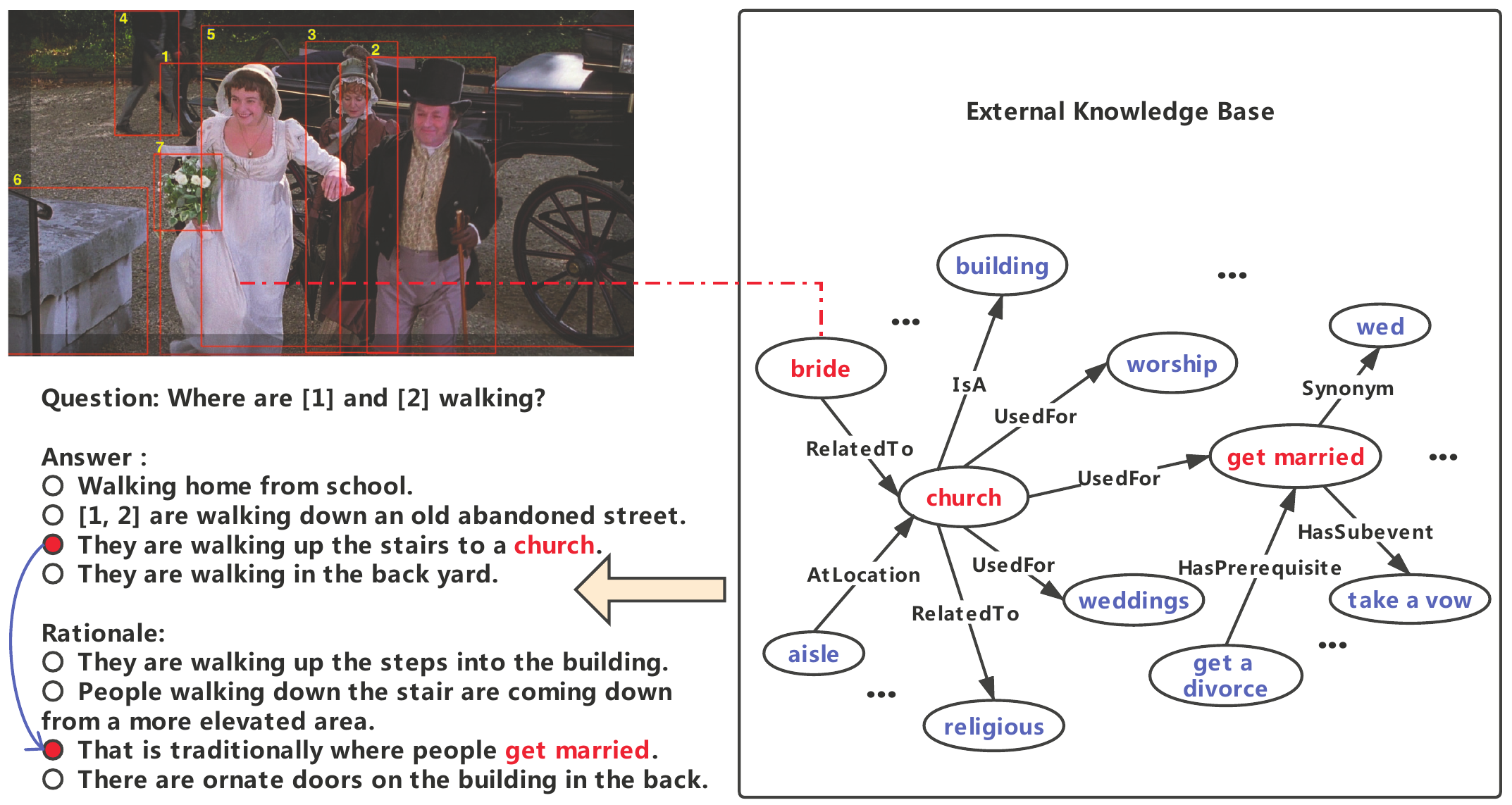}\\
  \caption{An illustrative example from the VCR benchmark (shown on the left). With the support of external commonsense knowledge (shown on the right), the question can be answered and reasoned more accurately.}
  \label{fig1}
\end{figure*}

\section{Introduction}
Recently, increasing attention has been focused on visual understanding, and great advances have been achieved in image caption (\cite{b1, b2, b3, b4, b5}) and visual question answer (VQA) (\cite{b6, b7, b8, b9}). Towards complete visual understanding, artificial intelligence models must perform cognition-level reasoning beyond recognition-level perception. To move towards this goal, the task of visual commonsense reasoning (VCR) \cite{b10} is proposed along with a well-devised new dataset. In VCR, given a challenging question about an image, a machine should answer it correctly and then provide a rationale justifying its answer. Besides detecting objects and their attributes, inferring the likely goals or reasons is needed.

In recent research, some task-specific models are proposed on the VCR task, such as R2C \cite{b10}, CCN \cite{b17} and HGL \cite{b18}, which achieve good results. The methods adopting the powerful BERT \cite{b11} model as the backbone for learning task-agnostic joint representation of image content and natural language, such as VisualBERT \cite{b12}, ViLBERT \cite{b13}, VL-BERT \cite{b14} and B2T2 \cite{b15}, have shown promising improvements on VCR. However, none of the existing methods have utilized commonsense knowledge in visual commonsense reasoning. In some cases, the explicit recognition results, such as objects or attributes, are not enough for accurate VCR. As not all of the required information is depicted in the image, we need the support of external knowledge to answer complex questions. Moreover, external knowledge supports cognitive reasoning, which is an essential challenge in the VCR task. As illustrated in Figure~\ref{fig1}, the left side of the figure describes an example from the VCR benchmark, the question could not be answered easily because there is no “church” shown in the figure. Based on the detected object “bride", only when the model is equipped with the commonsense knowledge “bride is related to church" and “church is used for getting married", the question could be answered and reasoned correctly.

Therefore, we incorporate commonsense knowledge into the cross-modal BERT, and propose a novel Knowledge Enhanced Visual-and-Linguistic BERT model in this paper. Specifically, to incorporate commonsense knowledge, we inject relevant entities extracted from ConceptNet \cite{b36} into the input sentence. In this way, the original sentence is transformed into a commonsense-knowledge-enriched sentence. Then, we propose a mechanism for sentence structure and semantic representation reservation. In order to keep the readability and structural information of the original sentence, we employ relative position embedding for the transformed sentence. Furthermore, inspired by \cite{b16}, to weaken the effect between the injected commonsense knowledge and other unrelated components in the input sequence,  we make the injected commonsense knowledge visible only to its related entity token, but not to other tokens in the original sentence or visual feature vectors via a visible matrix. We also adopt mask-self-attention mechanism to reserve the semantic and visual representations of the original input. Finally, we feed the token embedding of the commonsense-knowledge-enriched sentence, its special position embedding,  segment embedding, visual feature embedding, and the visible matrix to the pre-trained Visual-Linguistic BERT \cite{b14} for training and inference. 


Taking Figure~\ref{fig1} as an example, the object “bride" is the category label of a specific bounding box in the input image. When the model answers question based on the input image and text sequence (the input text sequence consists of question and one of the answers), it first retrieves the tokens contained in the input text sequence from the external knowledge base. For the token “church", the model could query its related entities as shown on the right of Figure ~\ref{fig1}. After the model injects the token “bride" from the external knowledge base into the original sentence, the representation of the token “church" is enriched by the injected token “bride". Then the attention score between the token “church" in the answer and the bounding box where “bride" is detected in the question will be high, which would help the model to choose the correct answer based on the original input and injected commonsense knowledge. 

We conduct comparative experiments on the VCR dataset. Compared to other task-specific models such as R2C \cite{b10}, CCN \cite{b17}, HGL \cite{b18}, and pre-trained task-agnostic multimodal BERT models such as VisualBERT \cite{b12}, ViLBERT \cite{b13}, Unicoder-VL \cite{b19}, B2T2 \cite{b15}, our KVL-BERT outperforms them by a large margin. To find the most effective way to integrate commonsense knowledge, besides our proposed KVL-BERT, we design and evaluate two variants: (1) Extract commonsense knowledge embedding corresponding to each token with transE \cite{bb16}, then input the word embedding and commonsense knowledge embedding to the multimodal BERT together. (2) Inject relevant entities extracted from ConceptNet into the input sentence in the same way as the KVL-BERT. Differently, we make the injected knowledge entity tokens share the same position embedding with their related token in the original sentence, and it lacks the mechanism of sentence structure and semantic representation reservation. In short, our contributions can be summarized as:
\begin{itemize}
\item We incorporate commonsense knowledge into the VCR task, and propose a novel KVL-BERT model. To the best of our knowledge, it is the first research to incorporate commonsense knowledge into the VCR task.
\item We design and evaluate three architectures of incorporating commonsense knowledge into the cross-modal BERT. The experimental results show that injecting commonsense knowledge into the input sentence with sentence structure and semantic representation reservation mechanism is the most effective way.

\item Compared to other task-specific models and general task-agnostic pre-training models, our KVL-BERT outperforms them by a large margin.

\end{itemize}

\section{Related Work}
\subsection{Visual commonsense reasoning}
As a critical step towards complete visual understanding, the task of visual commonsense reasoning (VCR) is proposed. Beyond recognition-level perception, the model must perform cognition-level reasoning. \cite{b10} introduces Recognition to Cognition Network (R2C) to model the necessary layered inferences for grounding, contextualization, and reasoning. \cite{b17} proposes a Cognition Connectivity Network (CCN)  including visual neuron connectivity, contextualized connectivity, and directional connectivity for reasoning. \cite{b18} proposes Heterogeneous Graph Learning (HGL) framework for seamlessly integrating the intra-graph and inter-graph reasoning in order to bridge the vision and language domain. Motivated by the success of BERT \cite{b11} in many natural language processing tasks, several researchers adopt BERT as the backbone for learning task-agnostic joint representation of image content and natural language, such as VisualBERT \cite{b12}, ViLBERT \cite{b13}, VL-BERT \cite{b14}, B2T2 \cite{b15}, Unicoder-VL \cite{b19} and UNITER \cite{b20}, which have shown promising improvements on VCR. However, none of the existing methods have utilized commonsense knowledge in visual commonsense reasoning, which we believe will be greatly helpful in this task. So we propose a novel model to incorporate commonsense knowledge into the cross-modal BERT.

\subsection{Pre-training for visual-linguistic tasks}

After the success of pre-training for computer vision (\cite{b21}, \cite{b22}) and natural language processing (\cite{b11}, \cite{b25}, \cite{b27}) tasks, a series of cross-modal pre-training models are designed. These models utilize self-supervised setting to get joint image-text embedding, gaining appealing results on various visual-linguistic tasks. Masked Language Model \cite{b11} and similar Masked Region Prediction \cite{b13} tasks are utilized in cross-modal pre-training. And similar to Next-Sentence Prediction \cite{b11}, Image-Text Matching (\cite{b13}, \cite{b14}, \cite{b20}) task in also widely used. \cite{b20} also adds extra scene graph prediction tasks (object prediction, attribute prediction and relationship prediction) in the pre-training phase, where the scene graph is constructed by parsing the text sentence into object nodes, attribute nodes and relationship nodes. These latest models are based on different variables of Transformers. VideoBERT \cite{b28} uses off-the-shelf networks to process video clips that are assigned to different clusters, whose ids will be predicted during pre-training. In ViLBERT \cite{b13}, LXMERT \cite{b29} and ERNIE-ViL \cite{yu2020ernie}, two-stream architecture is introduced. Two single-modal networks process the input image and sentence respectively, then a cross-modal Transformer combines two kinds of information. On the contrary, VisualBERT \cite{b12}, Unicoder-VL \cite{b19}, VL-BERT \cite{b14}, B2T2 \cite{b15} UNITER \cite{b20} and VILLA \cite{gan2020large} propose the single-stream architecture, where a single Transformer is applied to both image and text contents. Compared to the two-stream architecture, it fuses cross-modal information earlier and more flexibly. In our paper, we adopt the single-stream VL-BERT as the backbone to incorporate external commonsense knowledge.

\subsection{External knowledge integration}

Recent work has confirmed that the machine can become more powerful when incorporating external knowledge in many tasks, such as object detection (\cite{b30}, \cite{b43}), dialogue generation (\cite{b31}, {\cite{b44}}) and cloze style reading comprehension (\cite{b32}, \cite{b45}). \cite{b30} quantifies semantic consistency based on knowledge graphs and further re-optimizes object detection to achieve better consistency. The incorporation of commonsense knowledge promotes the dialogue generation system \cite{b31} to generate more accurate responses for both factoid-questions and knowledge grounded chats. By integrating knowledge, the model \cite{b32} can obtain more explicit evidence in the reading comprehension process. \cite{b16} solves the knowledge-driven problems in the plain text tasks leveraging domain-specific knowledge. In this paper, our goal is incorporating external commonsense knowledge into the visual commonsense reasoning task to answer complex questions even if the required information is not depicted in the image with cognitive reasoning.

\begin{figure*}[t]
  \centering
  \includegraphics[width=1.0\linewidth]{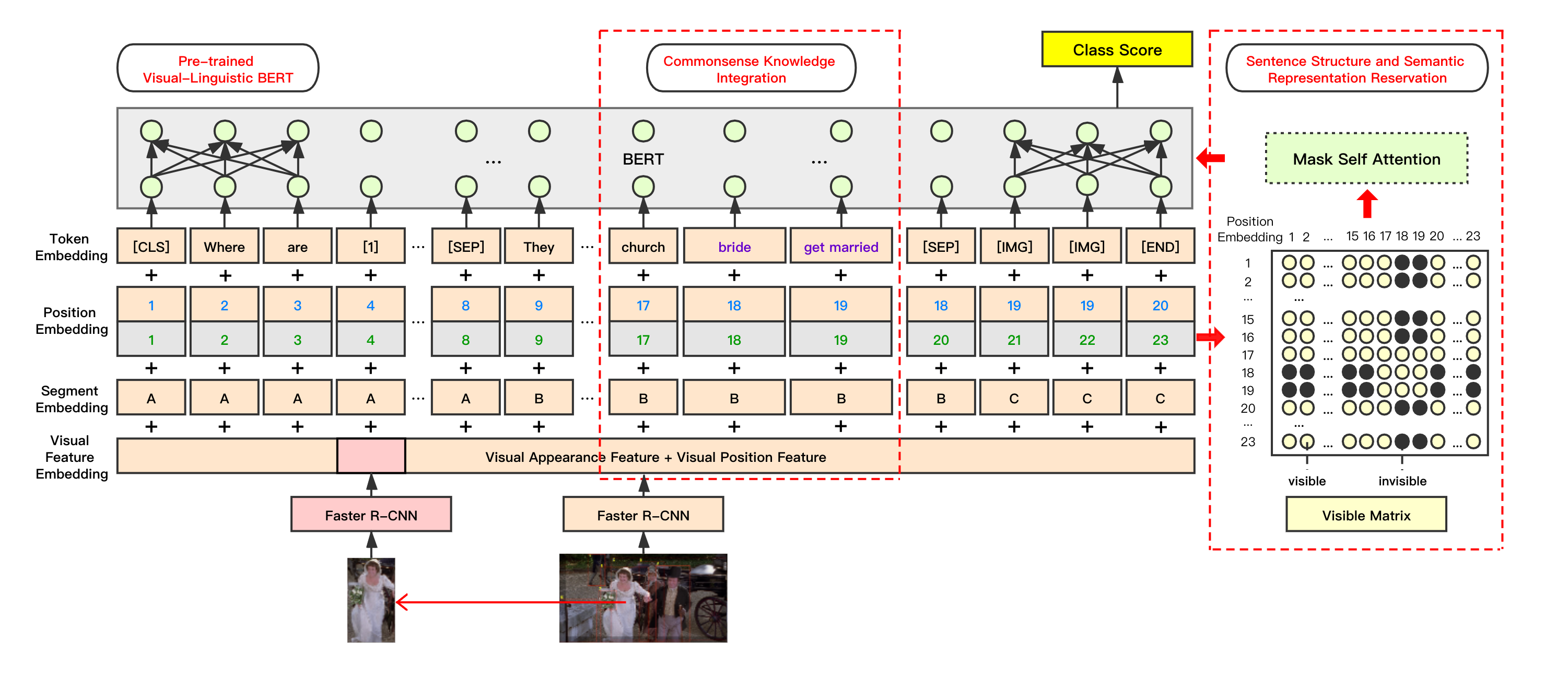}\\
  \caption{The architecture of KVL-BERT.}
  \label{fig2}
\end{figure*}

\section{Model Description}

Given an input image, the VCR task is divided into two subtasks: (1) $Q \rightarrow A$: given a question ($Q$), select  the correct answer ($A$) from candidate answers. (2) $QA \rightarrow R$: given a question ($Q$) and its correct answer ($A$), select the correct rationale ($R$) from candidate rationales. Both subtasks can be unified as choosing the correct response from candidate options given a query. For each query-response pair, the class score is calculated, and we choose the response with the highest score.


In this section, we present the overall framework of KVL-BERT and its detailed implementation, including the model architecture in Section \ref{A},  the method of commonsense knowledge integration in Section \ref{B}, the mechanism of sentence structure and semantic representation reservation in Section \ref{C}, and the pre-trained visual-linguistic BERT model in Section \ref{D}.

\subsection{Model architecture}
\label{A}



As shown in Figure~\ref{fig2}, the whole model architecture of KVL-BERT consists of three modules:

\emph{Commonsense knowledge integration} module is responsible to transform the original sentence into a commonsense-knowledge-enriched sentence. For an input sentence, this module retrieves relevant commonsense knowledge facts from ConceptNet and injects them into the original sentence. In Figure~\ref{fig2}, the purple tokens “bride" and “get married" are the injected commonsense knowledge for token “church". 

\emph{Sentence structure and semantic representation reservation} module is responsible to adjust the effect between the injected commonsense knowledge and other components in the original input. In Figure~\ref{fig2}, since the original absolute position indexes (marked in green) are changed due to the knowledge injection, we conduct relative position embedding (marked in blue) to keep the structural information of the original sentence. Then a visible matrix is constructed to limit the visible region of each token, which will be used to conduct mask-self-attention.


\emph{Pre-trained visual-Linguistic BERT} module is responsible to align tokens in the input sentence with regions in the input image, and learn a joint representation of visual and linguistic contents. In this module, besides all the components of BERT, visual feature embedding is introduced to model the input image. All the embeddings are then passed to the multi-layer Transformer to learn a new joint representation.

\subsection{Commonsense knowledge integration}\label{B}

We choose ConceptNet as the source of external commonsense knowledge, which is a knowledge graph that connects words and phrases of natural language with labeled and weighted edges. It can be seen as a large set of facts, and each fact $f_i$ is represented as a triple $f_i=(h, r, t)$, where $h$ and $t$  represent head and tail entities in the concept set $V$, $r$ is a relation type from the pre-defined set $R$, e.g., $\left([{dog}]_{h},[{HasA}]_{r},[{tail}]_{t}\right)$. 



Given an input sentence, we first retrieve the relevant commonsense knowledge facts via entity tokens contained in the input sentence. Each fact has a weight representing the credibility of it. The larger the weight is, the more credible the fact is. We sort the facts related to the input token by the weight value, because the facts with larger weight value are more trustworthy, i.e., they are more acceptable in the real world and more consistent with human cognition. Then we get the top $k$ commonsense knowledge entities from the sorted list and insert them after their relevant token ($k$ is a hyper parameter), while subsequent tokens in the sentence are moved backwards. In this way, the original sentence is transformed into a commonsense-knowledge-enriched sentence.



\subsection{Sentence structure and semantic representation reservation}\label{C}

The input sentence becomes unreadable and its structure is deformed by the injection of commonsense knowledge. To tackle this issue, we propose to conduct relative position embedding. In addition, we set a visible matrix and conduct mask-self-attention simultaneously to reserve the semantic and visual representations of the original input. 

\subsubsection{Relative position embedding}


For the self-attention mechanism in BERT, it does not take advantage of the position information of the word. In this case, even if two identical words appear in different positions, they will be encoded into a same vector when the model parameters are fixed. But in fact, these two same words appear in different positions may have different semantics, so the structural information of sentence will be utilized well by adding position embedding to the input of BERT. However, the position embedding is changed due to the injection of commonsense knowledge, which will deform the structure of the original sentence. To this end, we conduct relative position embedding for the commonsense-knowledge-enriched sentence. The position embedding of the original tokens is not changed, regardless of whether commonsense knowledge is injected, while the position embedding of the injected knowledge for a token increases from the position of the token. In this way, we can still use the structural information of the original sentence to calculate the self-attention score in the Transformer encoder. 

In addition, as the visual position information is expressed by its coordinate and size, we will take it into consideration during conducting visual feature embedding. Here we assign the same position embedding for all [IMG] tokens. 
\subsubsection{Visible matrix}

The injected commonsense knowledge will also change the representation of other components in the original input. Therefore, we set a visible matrix to weaken the effect between the injected commonsense knowledge and other unrelated components in the input sequence. For a certain token, the injected commonsense knowledge tokens are only related to it, but unrelated to other tokens contained in the original sentence, which are unrelated components. For example, in Figure~\ref{fig2}, for the token “church" in the input text sequence, the injected commonsense knowledge token “bride" is only related to the token “church", but unrelated to other tokens contained in the original input sentence, such as “walking" and “stairs". We suppose that the injected knowledge only acts on its related entity token and doesn't influence  other words or visual feature representation contained in the input sequence. Meanwhile, other words and visual feature representation shouldn't affect the representation of the external knowledge. For this reason, we set a visible matrix to limit the visible region of each token, i.e., we make the injected commonsense knowledge visible only to its related entity token, but not to other tokens in the original sentence or visual feature vectors. The visible matrix $W$ is defined as 
\begin{equation}
W_{i j}=\left\{\begin{array}{ll}
0, & w_{i} \text { is invisible to } w_{j} \\
1, & w_{i} \text { is visible to } w_{j}
\end{array}\right.
\label{eq1}
\end{equation}
where $w_i$ and $w_j$ are the $i^{th}$ and $j^{th}$ tokens in the commonsense-knowledge-enriched sentence, respectively. 
\subsubsection{Mask-self-attention}

Although we conduct relative position embedding to reserve structural information of the original sentence, another problem appears simultaneously: different tokens in the commonsense-knowledge-enriched sentence may share the same relative position embedding. When calculating self-attention score, these two unrelated tokens may obtain a high score because of the same position embedding. To preserve the semantic and visual representations of the original input, and weaken the effect between the injected commonsense knowledge and other unrelated components, we conduct mask-self-attention mechanism via the visible matrix, which could limit the self-attention area effectively. Formally, the mask-self-attention is described by 

\begin{equation}
{Q}^{t+1}, {K}^{t+1}, {V}^{t+1}=h^{t} {W}_{q}, h^t {W}_k, h^t {W}_v
\end{equation}

\begin{equation}
S^{t+1}={softmax}\left(\frac{Q^{t+1} K^{t+1^{\top}}+(W-1)* \mathrm{INF}}{\sqrt{d_{k}}}\right)
\end{equation}

\begin{equation}
h^{t+1}=S^{t+1} {V}^{t+1}
\end{equation}
where $h^{{t}}$ and $h^{t+1}$ denote the hidden state of the $t^{th}$ and $(t+1)^{th}$ mask-self-attention blocks, ${W}_q, {W}_k, {W}_v$ are trainable model parameters, and $Q^{t+1}$, $K^{t+1}$,  $V^{t+1}$ denote query, key and value respectively. ${W}$ is the visible matrix we defined in Eq.~\ref{eq1}. $\mathrm{INF}$ stands for an infinite number. $d_k$ is the scaling factor to counteract the effect of the dot products growing large in magnitude. $S^{t+1}$ denotes the attention score between query and key. In this way, if $w_j$ is invisible to $w_i$, $S^{t+1}_{ij}$ will approach 0 under the action of visible matrix, which means $w_j$ makes no contribution to the hidden state of $w_i$.

\subsection{Pre-trained visual-linguistic BERT}\label{D}
To extend the powerful pre-trained BERT model to visual-and-linguistic tasks, some researchers attempt to design cross-modal pre-training models, which can understand not only the semantic and visual contents, but the alignment and relationship between these two modals. In this paper, we adopt the pre-trained VL-BERT \cite{b14} as the backbone and incorporate external commonsense knowledge into it. 

In VL-BERT, two pre-training tasks are introduced. One is Masked Language Modeling with Visual Clues, which is similar to the Masked Language Modeling task utilized in BERT. The key difference is that visual clues are incorporated for capturing the dependencies among visual and linguistic contents. The model is trained to predict the masked words, based on the unmasked words and visual features. The other is Masked RoI Classification with Linguistic Clues, which is the dual task of the former. And the pre-training task is designed to predict the category label of the masked RoI from the other clues. Those pre-training tasks drive the network to not only model the dependencies in text and visual contents, but also to align the linguistic and visual contents.

Our KVL-BERT model takes token embedding, segment embedding, position embedding and visual feature embedding as the input into the pre-trained VL-BERT, these embeddings are then fed into a multi-layer Transformer to learn a cross-modal representation between visual regions and textual tokens. The details of the embeddings are as follows.

\subsubsection{Token embedding}
To encode the whole input text, first we merge the input query and one of the responses into a sentence separated by the special symbol [SEP]. Each token in this sentence is either a word or an explicit reference to the bounding box. We treat each word as the non-visual element and each explicit reference to the bounding box as the visual element respectively. For the visual elements, a special [IMG] token is assigned for each one of them. Following the standard text preprocessing method of BERT, we tokenize each input text into WordPieces \cite{b38}. The vocabulary is the same as BERT, which contains 30,522 tokens. 

\subsubsection{Segment embedding and position embedding}
The input elements from different sources are separated with three types of segments. For the subtask of $Q \rightarrow A$, question, answer, and RoIs (regions-of-interest) from the input image are separated into three different segments. While for the subtask of $QA \rightarrow R$, question with its correct answer, rationale, and RoIs from the input image are separated into three different segments. For position embedding, we adopt relative position embedding introduced in Section \ref{C}. 

\begin{table*}[htp]
\setlength{\abovecaptionskip}{0pt}
\setlength{\belowcaptionskip}{10pt}
\caption{Experimental results of our KVL-BERT model compared with other single models.}
\centering
\begin{tabular}{l|cccccc}
\hline
 & \multicolumn{2}{c}{$Q \rightarrow A$} & \multicolumn{2}{c}{$QA \rightarrow R$} & \multicolumn{2}{c}{$Q \rightarrow AR$} \\
 Model& val &test&val&test&val&test\\
  \hline
R2C \cite{b10} & 63.8 & 65.1 & 67.2 & 67.3& 43.1 &44.0\\  
CCN \cite{b17} &67.4 & 68.5 & 70.6 & 70.5 & 47.7 & 48.4\\
HGL \cite{b18} & 69.4&	70.1&	70.6&	70.8	&49.1&	49.8 \\

\hline
VisualBERT \cite{b12} &70.8&	71.6	&73.2&	73.2&	52.2	&52.4\\
ViLBERT \cite{b13} &72.4	&73.3&	74.5&	74.6	&54.0&	54.8\\
 Unicoder-VL \cite{b19} & 72.6	&73.4	&74.5&	74.4	&54.5&	54.9\\
B2T2 \cite{b15} & 71.9	&72.6	&76.0&	75.7	&54.9&	55.0\\
$\mathrm{UNITER}_{\mathrm{BASE}}$* \cite{b20} & 72.8 & - & 75.3 & - & 54.9 & - \\

\hline 
VL-$\mathrm{BERT}_{\mathrm{BASE}}$ \cite{b14} & 73.8 & - & 74.4 & - & 55.2 & - \\
KVL-$\mathrm{BERT}_{\mathrm{BASE}}$  (\textbf{ours}) &74.0 & - & 75.1 & - & 55.6& - \\
VL-$\mathrm{BERT}_{\mathrm{LARGE}}$ \cite{b14} & 75.5 & 75.8 & 77.9 &78.4 & 58.9 & 59.7 \\
KVL-$\mathrm{BERT}_{\mathrm{LARGE}}$  (\textbf{ours}) &\textbf{76.3}  & \textbf{76.4}&\textbf{78.6}  & \textbf{78.6} &\textbf{60.0}  & \textbf{60.3} \\
\hline
\end{tabular}
\label{tab2}
\end{table*}

\begin{table*}[htp]
\setlength{\abovecaptionskip}{0pt}
\setlength{\belowcaptionskip}{10pt}
\caption{Experimental results of ablation studies.}
\centering
\begin{tabular}{l|ccc}
\hline
Model& $Q \rightarrow A$ & $QA \rightarrow R$ & $Q \rightarrow AR$  \\
\hline
KVL-BERT w/o relative position embedding& 73.7 & 74.6& 55.0 \\
KVL-BERT w/o mask-self-attention& 73.3& 74.0 & 54.2 \\
KVL-BERT & \textbf{74.0}& \textbf{75.1}& \textbf{55.6} \\


\hline
\end{tabular}
\label{tab_ablation}
\end{table*}

\subsubsection{Visual feature embedding}
The visual feature embedding is a sum of visual appearance feature embedding and visual position feature embedding. The visual appearance feature embedding is extracted by Faster R-CNN \cite{b39}. For each visual element, its visual appearance feature is extracted on its reference bounding box. As for the non-visual element, its visual appearance feature is extracted on the whole input image. Additionally, to embed the position and size of a bounding box, each RoI is represented by a vector composed of normalized top-left and bottom-right coordinates  as $\left(\frac{{x}_{{LT}}}{{W}}, \frac{{y}_{{LT}}}{{H}}, \frac{{x}_{{RB}}}{{W}}, \frac{{y}_{{RB}}}{{H}}\right)$, where $\left({x}_{{LT}}, {y}_{{LT}}\right)$ and $\left({x}_{{RB}}, {y}_{{RB}}\right)$ denote the coordinate of the top-left and bottom-right corner, while ${H}$ and ${W}$ denote the height and width of the input image, respectively. Then, adopting  the method in \cite{b40}, the 4-D position vector is transformed into high-dimensional (under the same size of visual appearance feature embedding) visual position feature embedding.


\section{Experiments}

\subsection{Dataset and metrics}
We conduct experiments on the VCR \cite{b10} benchmark, a large-scale visual commonsense reasoning dataset containing over 212k (train set), 26k (validation set) and 25k (test set) questions on over 110k movie scenes. We follow this data partition in all of our experiments. 

The models are evaluated with classification accuracy in three modes: $Q \rightarrow A$ (given a question, select the correct answer from four candidate answers), $QA \rightarrow R$ (given a question and its correct answer, select the correct rationale from four candidate rationales), and $Q \rightarrow AR$ (given a question, select the correct answer first, then choose the correct rationale based on the answer). For the $Q \rightarrow AR$ mode, a sample will be treated as correct if and only if the model predicts both correct answer and correct rationale.

\subsection{Implementation details}
Our model adopts pre-trained parameters from the VL-BERT \cite{b14}, which are pre-trained jointly on Conceptual Captions \cite{b41} as visual-linguistic corpus, and BooksCorpus \cite{b42} and English Wikipedia as text-only corpus. The model is trained on the training set, and is evaluated on the validation and test sets. During training, we run our experiments on 4 NVIDIA Tesla V100 GPUs for 18 epochs, with the batch size of 256. The number of commonsense knowledge entities injected for each token is set to 2 (we will discuss it later). We use the SGD optimizer with base learning rate of 5e-3, momentum of 0.9, weight decay of 1e-4. Float16 operations are used to speed up the training process and reduce the usage of memory. 




\subsection{Quantitative evaluation}
We train and evaluate the models developed from the original $\mathrm{BERT}_{\mathrm{BASE}}$ and $\mathrm{BERT}_{\mathrm{LARGE}}$, where the subscripts “BASE" and “LARGE" are used to distinguish them. We compare our KVL-BERT with the VL-BERT  \cite{b14}. As shown in Table \ref{tab2}, our KVL-$\mathrm{BERT}_{\mathrm{BASE}}$ outperforms VL-$\mathrm{BERT}_{\mathrm{BASE}}$ on the validation set, and the KVL-$\mathrm{BERT}_{\mathrm{LARGE}}$ outperforms VL-$\mathrm{BERT}_{\mathrm{LARGE}}$ on the validation and test sets.


\begin{figure*}[htp]
  \centering
  \includegraphics[width=1\linewidth]{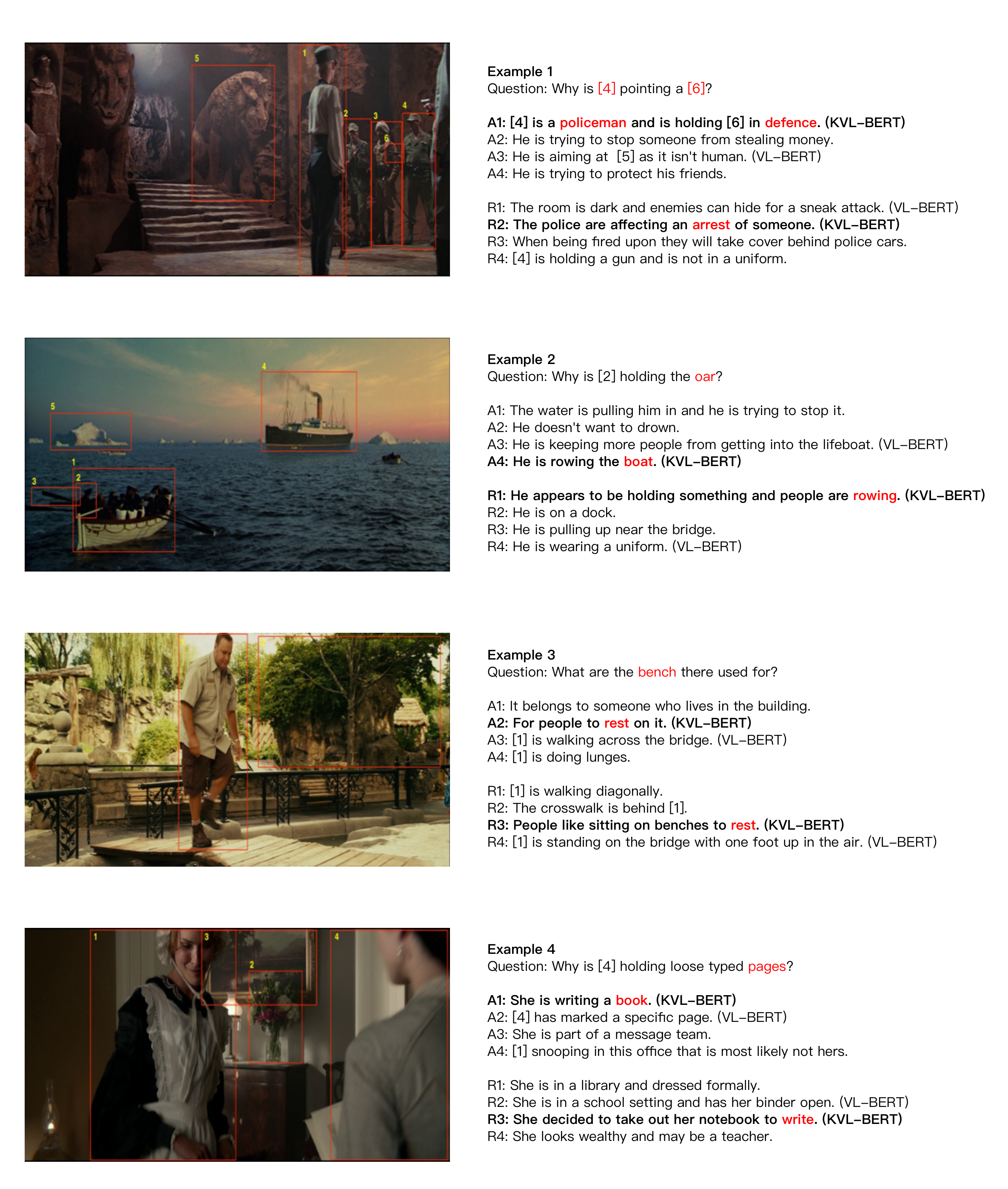}\\
  \caption{Examples of  $Q \rightarrow A$ and $QA \rightarrow R$ tasks from the VCR val set. The correct answer and rationale for each example is marked in bold. The answers picked by our KVL-BERT and baseline model VL-BERT are indicated in parenthesis. The tokens in red are the commonsense knowledge as the clue to answer and reason the question.}
  \label{fig_case}
\end{figure*}

Compared to other task-specific models such as R2C \cite{b10}, CCN \cite{b17}, HGL \cite{b18}, and existing pre-trained task-agnostic multimodal BERT models such as VisualBERT \cite{b12}, ViLBERT \cite{b13}, Unicoder-VL \cite{b19} and B2T2 \cite{b15}, our KVL-BERT outperforms these single models (not ensemble ones) by a large margin. 

In addition to the results listed in Table \ref{tab2}, some of the latest models have also achieved competitive results on the VCR task. $\mathrm{UNITER}_{\mathrm{LARGE}}$ \cite{b20} outperforms our KVL-$\mathrm{BERT}_{\mathrm{LARGE}}$ because it conducts two-stage pre-training: first pre-trains their model on task-agnostic pre-training datasets, and then pre-trains on the downstream task-specific dataset. VILLA \cite{gan2020large} performs large-scale adversarial training (task-agnostic adversarial pre-training and task-specific adversarial pre-training) based on UNITER \cite{b20}. ERNIE-ViL \cite{yu2020ernie} adds extra scene graph prediction tasks (object prediction, attribute prediction and relationship prediction) in the pre-training phase, where the scene graph is constructed by parsing the text sentence into object nodes, attribute nodes and relationship nodes. These three models outperform us due to the additional pre-training.


For the reason that pre-training is computationally expensive and time-consuming, we adopt the same comparison scheme as Unicoder-VL \cite{b19}, comparing our KVL-$\mathrm{BERT}_{\mathrm{BASE}}$ with the UNITER's one-stage pre-training model developed from the original $\mathrm{BERT}_{\mathrm{BASE}}$ model. It is denoted as $\mathrm{UNITER}_{\mathrm{BASE}}$* in Table \ref{tab2}, whose setting is similar to the our work. We directly use the results of $\mathrm{UNITER}_{\mathrm{BASE}}$* published in the UNITER paper \cite{b20}. As shown in Table \ref{tab2}, our KVL-$\mathrm{BERT}_{\mathrm{BASE}}$ outperforms $\mathrm{UNITER}_{\mathrm{BASE}}$* on the subtasks $Q \rightarrow A$ and $Q \rightarrow AR$, which strongly confirm the effectiveness of our commonsense knowledge incorporation method. 

Compared to the baseline VL-BERT model which extends pre-trained BERT to the visual-linguistic tasks, our KVL-BERT model outperforms it due to the incorporation of the commonsense knowledge. And we expect that introducing our proposed mechanism of incorporating commonsense knowledge into other pre-trained multi-modal BERT models will also bring improvement. In our future work, we will adopt more pre-training tasks to further improve our model.



\subsection{Case studies}\label{case study}
As shown in Figure~\ref{fig_case}, we show some examples to illustrate the effectiveness of our approach compared to the baseline model VL-BERT \cite{b14}.



Example 1 and Example 2 show how our model picks the right answers and rationales when the questions are about “why”. Based on the recognition-level perception such as detected objects and attributes, those reason-oriented questions can't be answered correctly. When the model is equipped with external commonsense knowledge, there would be enough clues supporting it to answer and reason the questions. In Example 1, when taking the question along with the first answer A1 as input, the related commonsense knowledge entity “gun" will be incorporated into the model through the token “policeman", so that the representation of the token “policeman" is enriched by the injected token “gun". Then the attention score between the token “policeman" in the answer and the bounding box where “gun" is detected in the question will be high. And when taking the question, the correct answer A1 and the rationale R2 as input, the related entity “policeman" will be incorporated into the model through the token “arrest", the representation of the token “arrest" is enriched by the external knowledge entity “policeman". Then the attention score between the token “arrest" in the rationale and the token “policeman" in the correct answer will be high. With the help of external commonsense knowledge, the model could answer and reason the question correctly. However, VL-BERT could not make the right choice as it is only equipped with the visual and text contexts, which are insufficient to answer and reason questions.

In Example 3, we show how our model answers the question about the function of the specific object. It is another kind of typical question that needs to be answered with the help of commonsense knowledge.

And there are also many examples similar to Example 4 that the objects or attributes in the input are ambiguous in the VCR dataset. In these situations, commonsense knowledge could provide extra semantic information to support answering and reasoning the questions.

In general, with the help of the external commonsense knowledge and our incorporation mechanism, the KVL-BERT could accurately choose the correct answer and rationale.


\subsection{Ablation study}\label{4.4}
We perform ablation studies to assess the impact of relative position embedding and mask-self-attention mechanism on the VCR val set with the model developed from the origin $\mathrm{BERT}_{\mathrm{BASE}}$. 

As shown in Table \ref{tab_ablation}, “KVL-BERT w/o relative position embedding" refers to conduct absolute position embedding, i.e., after inserting the external knowledge entities, the positions of all tokens in the overall transformed sentence are encoded in absolute sequence. “KVL-BERT w/o mask-self-attention" refers to remove the visible matrix from our model and just conduct self-attention mechanism. We can observed that without performing relative position embedding or mask-self-attention mechanism, the performance of the KVL-BERT declines. 

We infer that conducting absolute position embedding for the transformed sentence damages the structure information of the original sentence. And when visible matrix and mask-self-attention mechanism are not employed, i.e., all the tokens in the transformed sentence are visible to each other, injected external knowledge entities would bring knowledge noise for other tokens in the original input sentence. Those ablation studies prove the effectiveness of the relative position embedding and mask-self-attention mechanism.

\begin{table}[htp]
\setlength{\abovecaptionskip}{0pt}
\setlength{\belowcaptionskip}{10pt}
\caption{Experimental results of KVL-BERT and its two variants.}
\centering
\begin{tabular}{c|ccc}
\hline
Model& $Q \rightarrow A$ & $QA \rightarrow R$ & $Q \rightarrow AR$  \\
\hline
Variant \uppercase\expandafter{\romannumeral1}& 71.7 & 72.0& 51.6 \\
Variant \uppercase\expandafter{\romannumeral2}& 73.1 & 74.6 & 54.5 \\
KVL-BERT & \textbf{74.0}&\textbf{75.1} & \textbf{55.6} \\


\hline
\end{tabular}
\label{tab4}
\end{table}

\subsection{Variants and analysis}\label{4.5}
To find the most effective way to incorporate commonsense knowledge into the visual-and-linguistic BERT, we conduct the experiments with two variants of the KVL-BERT. We evaluate these three models on the validation set with the model developed from the original $\mathrm{BERT}_{\mathrm{BASE}}$.



For Variant \uppercase\expandafter{\romannumeral1}, we attempt to extract commonsense knowledge embedding corresponding to each token with transE. Given an input sentence, the model first retrieves the corresponding commonsense knowledge subgraph from ConceptNet for each token. The knowledge subgraph consists of a set of triples. Then the model conducts transE on the knowledge subgraph to get its embedding. Finally, the commonsense knowledge embedding is fed to the pre-trained VL-BERT \cite{b14} along with other embeddings. As shown in Table \ref{tab4}, the accuracy of Variant \uppercase\expandafter{\romannumeral1} is 2.3\%, 3.1\%, 4.0\% lower than the KVL-BERT on the subtasks $Q \rightarrow A$, $QA \rightarrow R$ and $Q \rightarrow AR$, respectively.

\begin{figure}[htp]
  \centering
  \includegraphics[width=1\linewidth]{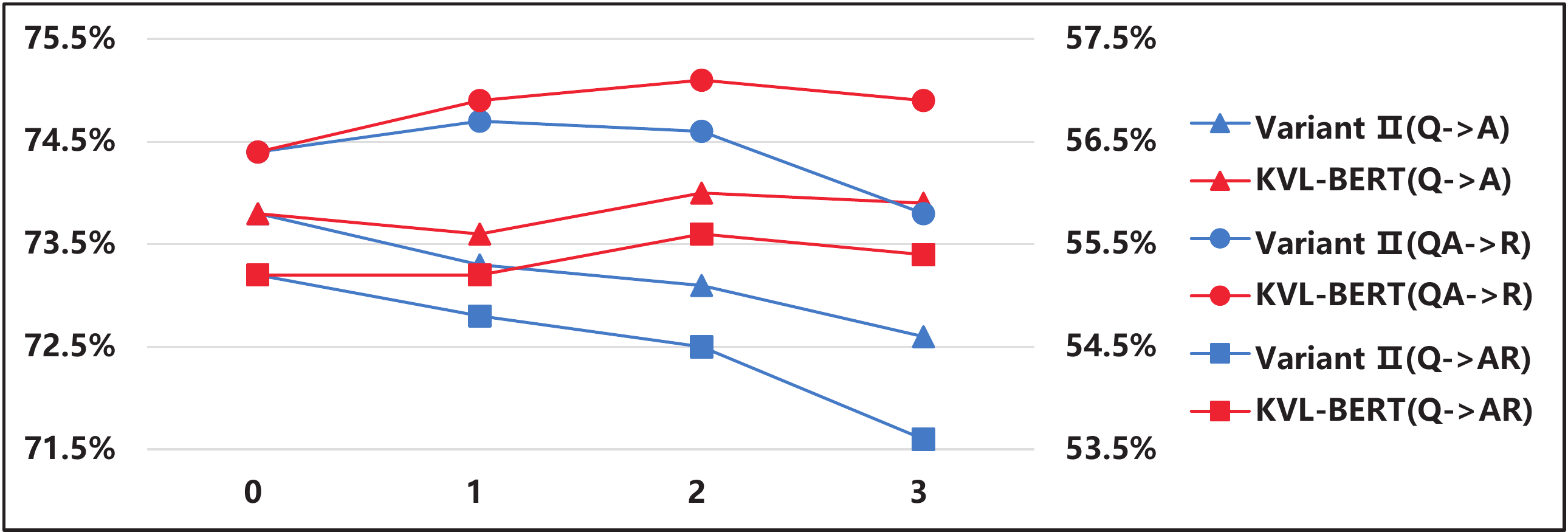}\\
  \caption{Experimental results for different numbers of knowledge entities injected for each token on KVL-BERT and Variant \uppercase\expandafter{\romannumeral2}. The scale of the left ordinate is used for the measurement of $Q \rightarrow A$ and $QA \rightarrow R$, while the right ordinate is used for $Q \rightarrow AR$.}

  \label{fig3}
\end{figure}

Variant \uppercase\expandafter{\romannumeral2} injects relevant entities extracted from ConceptNet into the input sentence in the same way as the KVL-BERT. Differently, we make the injected knowledge entity tokens share the same position embedding with their related token in the original sentence, and it lacks the mechanism of sentence structure and semantic representation reservation. As shown in Figure~\ref{fig3}, when the commonsense knowledge entities are injected, our KVL-BERT outperforms Variant \uppercase\expandafter{\romannumeral2} on all the subtasks in VCR, which verifies the effectiveness of sentence structure and semantic representation reservation mechanism. Note that these two models are identical when the number of knowledge entities injected for each token equals zero, i.e., there is no commonsense knowledge incorporated into the original sentence. The results listed in Table \ref{tab4} is the best performance of KVL-BERT and Variant \uppercase\expandafter{\romannumeral2}.

In addition, as shown in Figure~\ref{fig3}, the accuracy rate of Variant \uppercase\expandafter{\romannumeral2} generally decreases as the number of  knowledge entities injected for each token increases. On the contrary, this issue does not appear in our KVL-BERT model, which credits to sentence structure and semantic representation reservation mechanism. Note that the KVL-BERT achieves the best performance when the number of commonsense knowledge entities injected for each token equals 2. When it increases to 3, the classification accuracy decreases, we infer that some knowledge noise is incorporated in this situation.




\section{Conclusion}
In this paper, we propose a novel KVL-BERT model to incorporate commonsense knowledge into the visual-and-linguistic BERT, which can improve the cognition-level visual understanding and reasoning abilities. Besides taking visual and linguistic contents as input, external commonsense knowledge extracted from ConceptNet is integrated into the multi-layer Transformer. In order to reserve the structural information and semantic representation of the original sentence, we propose conducting relative position embedding and mask-self-attention to weaken the effect between the injected commonsense knowledge and other unrelated components in the input sequence. In addition, to find the most effective way to integrate commonsense knowledge, we design and evaluate two variants of the KVL-BERT. When applying on the visual commonsense reasoning task, compared to other task-specific models and general task-agnostic pre-training models, our KVL-BERT outperforms them by a large margin. We will apply our KVL-BERT model to more tasks of visual sense analysis and interpretation for future research.

\section*{Acknowledgment}

This work was supported by National Key Research and Development Program of China [Grant No. 2020AAA0106600]; and National Natural Science Foundation of China [Grant Nos. 61976021 and U1811262].

\bibliographystyle{elsarticle-num}
\bibliography{mybib}

\end{document}